\documentclass[runningheads]{article}
\date{}
\usepackage{url}
\usepackage{times}
\usepackage{graphicx}
\usepackage{mathtools}
\usepackage{amsmath,amsfonts}
\usepackage{amssymb}
\usepackage[ruled, linesnumbered, vlined]{algorithm2e}
\usepackage{todonotes}
\usepackage{xspace}

\usepackage{amsmath}
\usepackage{amsthm}
\usepackage{booktabs}
\usepackage{multirow}
\usepackage{algorithmicx}
\usepackage{algpseudocode}
\usepackage{enumitem}
\usepackage{times}
\usepackage{xparse}
\usepackage{mathtools}
\usepackage{amssymb}
\newtheorem{theorem}{Theorem}
\newtheorem{example}[theorem]{Example} 

\def\rConstrHelper(#1,#2,#3,#4){{#1}\ \substack{#2 \\ #3}\ {#4}}

\newcommand{\sss}{{\mathrel{\kern.25em{\sqsubseteq}\kern-.5em \mbox{{\scriptsize *}}\kern.25em}}}
\newcommand{\smallsss}{{\mathrel{\kern.25em{\sqsubseteq}\kern-.4em \mbox{{\scriptsize *}}\kern.25em}}}

\newlength{\indxlength}
\begin{document}
\title{Rule Learning as Machine Translation using the Atomic Knowledge Bank  \thanks{Supported by the Norwegian Research Council grant number 316022.}}
%
%
\author{Kristoffer Æsøy and
Ana Ozaki
\\
University of Bergen \& University of Oslo, Norway  \\
\texttt{kristoffer.asoy@student.uib.no, anaoz@uio.no}}
%
%
%
\maketitle              
\begin{abstract}
Machine learning models, and in particular language models, are being applied to various tasks that require reasoning.
While such models are good at capturing patterns their ability to reason in a trustable and controlled manner is frequently questioned.
 On the other hand, 
logic-based rule systems allow for controlled 
inspection and already established verification methods. However it is well-known that creating such systems manually is time-consuming and prone to errors.
We explore the
  capability of transformers to translate sentences expressing rules in natural language into logical rules. 
   We see reasoners as the most reliable tools for performing logical reasoning and focus on translating language into the format expected by such tools. 
We perform experiments using the DKET dataset from the literature and create a dataset 
for language to logic translation based on the Atomic knowledge bank.
\end{abstract}
\section{Introduction}

Language models are being applied in various tasks 
that require reasoning. While these models are good at capturing patterns, despite recent improvements, they still fail to perform complex logic-based reasoning~\cite{DBLP:journals/corr/abs-2303-14310}. Such models can hallucinate, providing wrong/inconsistent information to users as if   correct.
As it happens with any kind of black-box machine learning model, ethical   issues are also of concern
and these issues are extremely hard to fix~\cite{10.5555/3379082}.
One approach to deal with these challenges is to have additional methods 
for checking the 
responses of   language models, acting as filters. 

In this work we explore the idea of converting sentences into logic, so that 
  we can perform consistency checks using logic reasoners  developed by the automated reasoning community. 
By taking the capability of language models to deal with language variability while leaving (at least part of) the reasoning task with dedicated, provably correct tools based on deductive reasoning, we can
exploit the best of both worlds.
  There are multiple challenges
associated with translating
expressions in natural language into logic.
One of them is that there are many expressions that cannot be translated into  logic in a natural way. 

 Even if we restrict to statements that
have a natural translation, a main challenge is to train 
  language models for translating
  from  (controlled) natural language into logic, due to 
  the lack of datasets where the examples are pairs $(s,t)$ with $s$ being a statement and $t$ its translation into logic. 
 A seminal approach in this direction has been developed by Petrucci et al.~\cite{PETRUCCI201866}. The authors aim
 at translating definitory sentences into 
 concept inclusions in the $\mathcal{ALCQ}$ ontology language.   
They create the DKET dataset (which stands for \emph{D}eep \emph{K}nowledge \emph{E}xtraction from \emph{T}ext) and 
perform experiments using recurrent neural networks (RNNs).
Some other recent approaches for translating from natural language into ontologies using Transformers include the work by Korel et al.~\cite{computers12010014} and the work by Wachowiak et al.~\cite{wachowiak_et_al:OASIcs.LDK.2021.22}.
In this work, we explore the capability of language models based on the {Transformer} architecture for the task of translating rules in natural language into \emph{first-order rules}. 
Converting sentences into rules can be useful to verify 
the output of language models, creating a filter that checks for the consistency of 
the responses. We focus on the Atomic knowledge bank~\cite{atomic2019} which contains a large number of common-sense \textbf{if-then} statements.



 
\paragraph{Contribution} 
To the best of our knowledge this is the first work that explores the capability of Transformers to translate from sentences expressing common-sense rules into first-order logic rules.  
We address the lack of datasets for performing such translations and create the Atomic dataset, using   the Atomic knowledge bank.
After creating the Atomic dataset,
we perform experiments using Transformers. We also run experiments using the DKET dataset~\cite{PETRUCCI201866}, originally tested on RNNs (instead of Transformers), and make a comparison. 
Our results indicate that Transformers require more training data than the RNN approach by~\cite{PETRUCCI201866} but once provided with more data it performs better.  


\section{Rule Learning as a Machine Translation Task}

We briefly describe our rule learning approach based on  translations from (controlled) natural language.
The source of the translation are 
statements from the Atomic knowledge bank~\cite{atomic2019}
that express rules, as we can see in this  example.
\begin{center}
``If PersonX paints the portrait of PersonY then PersonX is creative.''
\end{center}

The target of the translation is a
rule in first-order logic. In detail,
a \textit{rule in first-order logic} is an expression of the form:
\[\forall \boldsymbol{x},\boldsymbol{y}.(\varphi[\boldsymbol{x},\boldsymbol{y}] \rightarrow \exists\boldsymbol{z}.\psi[\boldsymbol{y},\boldsymbol{z}])\]
where the body $\varphi$ and the head $\psi$ are conjunctions of atoms, and $\psi$ contains at least one
conjunct.
The rule in natural language given   above can be expressed in first-order logic using the following formula:
\[\forall x y_1 y_2.((\mathsf{person}(x) \wedge \mathsf{paints}(x,y_1,y_2) \wedge \mathsf{portrait}(y_1)\wedge\mathsf{person}(y_2)) \rightarrow \mathsf{creative}(x)).\]


We use transformers to make the translation between rules in
natural language and rules formulated in logic. However, even if we only consider  statements that
have a natural translation, such as the expressions found in the Atomic knowledge bank and definitory sentences~\cite{PETRUCCI201866}, there are not
many datasets available. In the next section,
we describe how we created the Atomic dataset, which contains pairs of the form $(s,t)$ where $s$ is a statement in the Atomic knowledge bank
and $t$ is its corresponding translation as a rule in first-order logic.



\section{The Atomic Dataset and   DKET}\label{sec:atomic}
In this section we first describe  the Atomic knowledge bank and how the Atomic dataset (with the translations into first-order logic rules) has been created for this work. We then describe the DKET dataset previously proposed in the literature to perform translations from definitory sentences into $\mathcal{ALCQ}$ using RNNs~\cite{PETRUCCI201866}.

\subsection{The Atomic Knowledge Bank}\label{atomic}
Atomic~\cite{atomic2019} is a massive atlas consisting of 877k descriptions of inferential knowledge. The KB focuses on the everyday commonsense understanding of rules, organized as typed if-then relations.
The data in Atomic was collected using a crowdsourcing framework,
where responders are asked to write answers to specific events. In other words, \textit{if} a certain event happened, \textit{then} a question is asked to infer something about that event. This could be what would necessarily precede the event to make it possible, what the event says about the person performing the event, or what could follow as a result of the event. 

For example, in the statement given in the previous section, stating that the ``person is creative'' corresponds to what the event says about the person performing the event (of painting a portrait).
All of these aspects relating to the event, which can be asked questions about, are known in the KB as \textit{inferential dimensions}, listed in Table \ref{table:infdims}.

\begin{table}[t]
    \begin{tabular}{c|c|c}
        Inf. dim. & Question & Category \\
        \hline
        xIntent & What does PersonX \textbf{intend} to do? & Mental-State \\
        xReact & How would PersonX typically \textbf{react} after this event? & Mental-State \\
        xNeed & Does PersonX \textbf{need} to do anything before this event? & Event \\
        xWant & What does PersonX likely \textbf{want} to do next after this event? & Event \\
        xEffect & What \textbf{effect} would this event typically have on PersonX? & Event\\ 
        xAttr & What characteristic would you \textbf{attribute} to PersonX? & Persona \\
        oReact & How would others typically \textbf{react} after this event? & Mental-State \\
        oWant & What would others typically \textbf{want} after this event? & Event \\
        oEffect & What \textbf{effect} will this event typically have on others? & Event \\
    \end{tabular}
    \vspace{0.2cm}
    \caption{The inferential dimensions of Atomic and their corresponding categories}
    \label{table:infdims}
\end{table}

One of the fundamental features of the Atomic KB is the crowd-sourced nature of the inferences in the \textit{typed if-then} relations. This allows the responder to express themselves however they want, and thus more accurately reflects the natural language of responses than simply restricting the user to only use certain words or sentence lengths. Inherently, this is not an issue, but there is an overall issue with what seems to be a lack of postprocessing performed on the inferences collected. In the original work~\cite{atomic2019} there is no indication of such a process, nor can it be observed from working with the data as we find certain elements expressed in many ways.
A difficulty arises if one wishes to capture, or analyze every instance of the individuals involved in the sentences. In the events provided by the framework, we find that they are always referred to as PersonX and PersonY with these exact spellings. This is not the case when we look at the inferences, where the respondees have found a multitude of ways to refer to them. 

\begin{example}
Here are some of the examples of expressions
  in  Atomic:\\
    PersonX builds PersonX's houses. $\rightarrow$ \textbf{X's} family has a new home to live in. \\
    PersonX replaces PersonY's tire. $\rightarrow$ to pay \textbf{person x}. \\
    PersonX provides PersonY description. $\rightarrow$ to help \textbf{him}.
\end{example}

Observing the few examples above we can immediately see that the issue is encountered in many different ways. In the first case, we notice that the ``Person'' part of ``PersonX'' has been omitted by the responder, which can be considered quite natural as we humans still infer quite easily which individual we are referring to as the distinctive part is still mentioned. We can see another example of the same issue handled differently in the second inference, where instead the ``Person'' and ``X'' parts have been separated by a space. Again, quite a natural way of expressing anonymous individuals seen in other literature and it still stays unambiguous to who they refer to. In both of these cases, the problem boils down to pattern matching, as we can still capture them as instances of ``PersonX'' by looking for instances where ``X'' is isolated and where it follows after the word ``person'' in the inferences. They merely introduce the necessity to perform more exhaustive searches since there are more patterns to look for.

The last   example on the other hand creates different, less easily solvable issues. We can see that the gender of ``PersonY'' has been assumed and is just referred to as ``him''. This introduces the problem of us having to infer which individual these terms are assigned to in each instance individually. In this example, it is quite clear, but there is no guarantee that we can correctly identify who a pronoun is describing in all inferences. In particular, in the types where we ask how it relates to \emph{others} in a given event such as in \verb|oEffect| and \verb|oReact| seen in Table \ref{table:infdims}. Atomic refers sometimes to \textit{implicit} individuals involved in an event\cite{atomic2019}, which means that sometimes we also need to make assumptions  if someone is referred   ambiguously in the inference, if this is one of the named individuals of the event or a potential third party. In addition, we wish to automate the process of creating the formula for each \textbf{if-then} expression, and would therefore somehow find some consistent way of handling them. We consider who the pronouns describe to be determined as a result of the appearance of PersonX and PersonY in the event and what type of inferential dimension we are dealing with.

\paragraph{Removal of PersonZ}
The Atomic KB concerns itself with the individual PersonX, often how it would be affected by events but also sometimes in relation to others. In most cases the other is the individual ``PersonY'', and sometimes it is simply implied. Though, in a few cases a ``PersonZ'' is also included, when PersonX is related to more than one other individual for the event. However, they are exceptionally rare and introduce weird situations that are not easy to model. 
For example:
    ``PersonX takes PersonY in PersonZ arms.'', 
    ``PersonX puts PersonY end to PersonZ.''  
    ``PersonX invites PersonZ's friend PersonY.''     
As seen in the examples, it is not always    clear what the intention is when PersonZ is involved. It also creates an even stronger uncertainty in terms of which individual the inference references when we ask about how the event relates to others. In most cases, we already need to create an assumption that when PersonY is involved that they are considered the ``other'', but with an additional individual this becomes even more dubious. Thus, we have decided to remove all events where PersonZ appear, which were a total of less than $1000$ total.

\subsection{Creating the Atomic Dataset}
We   present the algorithms used
to create the Atomic dataset translating from semi-structured sentences into first-order logic.
The algorithms work on each 
if-then statement individually. Firstly, it splits it into three parts, the event, the inferential dimension, and the inference. 

\begin{example}
Consider the sentence: ``If PersonX paints the portrait of PersonY then PersonX is creative.''
 In this sentence, the \textbf{event} 
 is that \emph{a person paints the portrait of another (person)}.
 The \textbf{inferential dimension} is
 \emph{PersonX}.
 Finally, the \textbf{inference} is
 \emph{(to be) creative}.
\end{example}

\SetKwInput{KwData}{input}
\SetKwInput{KwResult}{output}
\begin{algorithm}[H]
\label{alg:Atomic_to_Rule}
\SetAlgoVlined
\LinesNumbered
\caption{Atomic to Rule}
\KwData{Two lists of pairs $(w,t)$ called $\mathsf{event}$ and $\mathsf{inference}$, where $w$ is a word and ${t}$ is a         POS-tag,
        a word that is the $\mathsf{inferential\_dimension}$,
        and a boolean $\mathsf{add\_quantifiers}$ to decide whether to quantify the variables or not.\\}
\KwResult{A FOL-rule for the Atomic if-then statement.}
\BlankLine
$\mathsf{body, body\_vars := EventToBody}(\mathsf{event})$\\
$\mathsf{head, head\_vars := InferenceToHead}( \mathsf{inference, body, inferential\_dimension})$\\
\eIf{$\mathsf{add\_quantifiers}$}
    {$\mathsf{rule := \forall body\_vars} (\mathsf{body \rightarrow \exists head\_vars} (\mathsf{head}))$}
    {$\mathsf{rule := body \rightarrow head}$}
\textbf{return} $\mathsf{rule}$
\end{algorithm}

Algorithm~\ref{alg:Atomic_to_Rule}   constructs the logical rule by calling
Algorithm~\ref{alg:Event_to_Body} for extracting the body of the rule from the event and 
Algorithm~\ref{alg:Inf_to_Head} for extracting the head of the rule from the inference.
In Algorithm~\ref{alg:Event_to_Body}, we try to capture three separate types of atoms, the individuals, the verb expression, and the object expression (if it exists). To do this we first identify all the words that have the \verb|IND| tag in the event, which represents the individuals, and remove them from the event. Then, we try to separate the verb expression and the potential object. 
Algorithm~\ref{alg:Inf_to_Head} is a bit more complicated as
it might encounter more than a single object due to the more freeform structure provided by the respondees (Section~\ref{atomic}).
We also have to take in account   the fact that some inferential dimensions do not concern themselves with PersonX, but rather how \textit{others} are affected by the event. This affects which variables become connected to the atoms.

The algorithms   here are not able to make a perfect translation of all sentences in the Atomic knowledge bank. The main limitations that we found were: (1) a small number of statements expressed disjunction (e.g. ``...wants to talk with a family member or a friend''), which is not in the scope of  first-order logic rules (our target language); (2) variations in the sentence when referring to the same entity.
For example, ``If PersonX paints PersonX's portrait then PersonX wants to hang the painting'' was translated into 
\[\forall xz(\mathsf{person}(x) \wedge \mathsf{paints}(x, z) \wedge \mathsf{portrait}(z) \rightarrow \exists a 
(\mathsf{toHang}(x, a) \wedge \mathsf{painting}(a))).\]
In this case the connection between portrait and painting was lost.
Nevertheless the algorithms could generate a large number of correct translations that allowed us to test the ability of Transformers to translate \textbf{if-then} statements
into rules in first-order logic. 

\SetKwInput{KwData}{input}
\SetKwInput{KwResult}{output}
\begin{algorithm}[t]
\label{alg:Event_to_Body}
\SetAlgoVlined
\LinesNumbered
\caption{Event to Body}
\KwData{List of pairs $(w,t)$ called $\mathsf{event}$, where $w$ is a word and $t$ is a POS-tag.}
\KwResult{The body of the Atomic rule and a list of variables found in the body.}
\BlankLine
$\mathsf{verb =obj =body =variables} :=\emptyset$\\
$\mathsf{verb\_finished := False}$\\
Remove $(\mathsf{PersonX,IND})$ and $(\mathsf{PersonY,IND})$ (if it exists) from the $\mathsf{event}$ list, and add them to $\mathsf{body}$.\\
Add $x$, $z$ to $\mathsf{variables}$ and also $y$, if $\mathsf{PersonY}$ occurs in $\mathsf{event}$.\\
\For{$(w, t) \in \mathsf{event}$}{
    \If{$\mathsf{verb} \neq \emptyset \text{ \upshape{and} } t \in [\mathsf{JJ, NN, NNS}]$}
        {$\mathsf{verb\_finished := True}$}
    \eIf{$\mathsf{verb\_finished}$}
        {Add $w$ to $\mathsf{obj}$.}
        {Add $w$ to $\mathsf{verb}$.}
}
Let $v$ be the concatenation of the words in $\mathsf{verb}$.\\
\eIf{$(\mathsf{PersonY,IND}) \in$ $\mathsf{event}$}
    {Add $v(x, z, y)$ to $\mathsf{body}$.}
    {Add $v(x, z)$ to $\mathsf{body}$.}
\If{$\mathsf{obj} \neq \emptyset$}
    {Let $o$ be the concatenation of the words in $\mathsf{obj}$ and
     add $o(z)$ to $\mathsf{body}$.}
\textbf{return} $(\mathsf{body,variables})$
\end{algorithm}

\SetKwInput{KwData}{input}
\SetKwInput{KwResult}{output}
\begin{algorithm}[H]
\label{alg:Inf_to_Head}
\SetAlgoVlined
\LinesNumbered
\caption{Inference to Head}
\KwData{List of pairs $(w,t)$ called $\mathsf{inference}$, where $w$ is a word and $t$ is a POS-tag,
        the $\mathsf{body}$ as a list of atoms and a word that is the $\mathsf{inferential\_dimension}$.}
\KwResult{The head of the Atomic rule and a list of variables  in the head.}
\BlankLine
$\mathsf{verb =cur\_obj =objects =head =variables} :=\emptyset$\\
$\mathsf{verb\_finished := False}$\\
Remove $\mathsf{PersonX}$ and $\mathsf{PersonY}$ from the $\mathsf{inference}$, and add them to $\mathsf{head}$ if they do not occur in the $\mathsf{body}$.\\
Add $x$ and $y$ to $\mathsf{variables}$ if $\mathsf{PersonX}$ and $\mathsf{PersonY}$ appear in the $\mathsf{inference}$ but are not already in the $\mathsf{body}$.\\
\For{$(w, t) \in \mathsf{inference}$}{
    \If{$\mathsf{verb} \neq \emptyset \text{ \upshape{and} } t \in [\mathsf{CC, DT, PRP, PRP}\$]$}
        {\eIf{$\mathsf{verb\_finished = True}$}
            {Add $\mathsf{cur\_obj}$ to $\mathsf{objects}$.\\
             Add $\mathsf{variable}$ for $\mathsf{cur\_obj}$ to $\mathsf{variables}$.\\
             $\mathsf{cur\_obj} := \emptyset$.}
            {$\mathsf{verb\_finished := True}$}
        }
    \eIf{$\mathsf{verb\_finished}$}
        {Add $w$ to $\mathsf{cur\_obj}$.}
        {Add $w$ to $\mathsf{verb}$.}
}
\If{$\mathsf{cur\_obj} \neq \emptyset$}
    {Add $\mathsf{cur\_obj}$ to $\mathsf{head}$.\\
    Add variable for $\mathsf{cur\_obj}$ to $\mathsf{variables}$.}
\eIf{$\mathsf{inferential\_dimesion = PersonX}$}
    {$\mathsf{subject} := x$\\
    \eIf{$\mathsf{PersonY} \text{ \upshape{occurs in} } \mathsf{body} \text{ \upshape{or} } \mathsf{head}$}
        {$\mathsf{target} := y$}
        {$\mathsf{target := none}$}
    }
    {\eIf{$\mathsf{PersonY} \text{ \upshape{occurs in} } \mathsf{body} \text{ \upshape{or} } \mathsf{head}$}
        {$\mathsf{subject} := y$}
        {$\mathsf{subject} := u$}
     $\mathsf{target} := x$}
\end{algorithm}

\begin{algorithm}[H]
\setcounter{AlgoLine}{30}
    Let $v$ be the concatenation of the words in $\mathsf{verb}$.\\
    \eIf{$\mathsf{objects} \neq \emptyset$}
        {\eIf{$|\mathsf{objects}| = 1$
        \upshape{and equals the object atom in the} $\mathsf{body}$}
            {\eIf{$\mathsf{target} \neq \mathsf{none}$}
                {Add $v(\mathsf{subject}, z, \mathsf{target})$ to $\mathsf{head}$.}
                {Add $v(\mathsf{subject}, z)$ to $\mathsf{head}$.}
            }
            {\For{$\mathsf{obj} \in \mathsf{objects}$}
                {Let $\mathsf{o\_var}$ be the variable for $\mathsf{obj}$ in variables.\\
                Let $o$ be the concatenation of the words in $\mathsf{obj}$.\\
                \eIf{$\mathsf{target \neq none}$}
                    {Add $v(\mathsf{subject}, o\_var, \mathsf{target})$ to $\mathsf{head}$.}
                    {Add $v(\mathsf{subject}, \mathsf{o\_var})$ to $\mathsf{head}$.}
                Add $o(\mathsf{o\_var})$ to $\mathsf{head}$.
                }
            }
        }
        {\eIf{$\mathsf{target \neq none}$}
            {Add $v(\mathsf{subject, target})$ to $\mathsf{head}$.}
            {Add $v(\mathsf{subject})$ to $\mathsf{head}$.}
        }
    \textbf{return} $(\mathsf{head,variables})$
\end{algorithm}

\subsection{The DKET Dataset}

The Deep Knowledge Extraction from Text (DKET) project from the literature~\cite{PETRUCCI201866} tests the use of a RNN model to translate natural language definitions into the description logic language $\mathcal{ALCQ}$. The model they created was designed to purely look at the syntactic structure of the definitions, and create ontologies by either copying words from the definition or emit a logical symbol defined in $\mathcal{ALCQ}$. The result of this was that the dataset used to perform the translation did not need to contain any semantic meaning, just match the syntax.

Such a dataset did not exist beforehand, and their approach to create it was to design a handcrafted context-free grammar to create definitions with corresponding  translations into $\mathcal{ALCQ}$ concept inclusions. The $158$ production rules allow for a large  syntactic variation that is guaranteed to be correct. However, the grammar only creates templates for sentence structures where the content words, the words that convey the meaning of the sentence, have been anonymized such that a sentence structure can be used again and again to create new sentences. In other words, only the \textit{definiendum}, \textit{genus proximum}, and any \textit{differentia specifica} are needed to be filled in from a vocabulary. The vocabulary used for their dataset consists of $2841$ nouns, $1629$ adjectives, and $897$ verbs~\cite{PETRUCCI201866}.

\section{Experiments}
We perform experiments on neural machine translation using the Transformer
language model.
In our experiments, we employ the DKET dataset from the literature~\cite{PETRUCCI201866} and the Atomic dataset 
created in this work (see Section~\ref{sec:atomic}).
The code is available on Github\footnote{\url{https://github.com/KrisAesoey/AtomicTranslation}}.
 To evaluate how well the Transformer performs the translations we use three metrics that were used on the DKET experiments from the literature~\cite{PETRUCCI201866}. 
 \textbf{Average per-formula accuracy} (FA), which tells us how many of the rules are perfectly translated, \textbf{average edit distance} (ED), where we count the number of operations necessary to correct the predicted rule, and finally \textbf{average per-token accuracy} (TA) which measures how many of the tokens of each prediction are at the correct index as a percentage score.
 
 \subsection{DKET using RNN and Transformer}

In Table~\ref{tab:dket_results} we can see the results that were achieved in the reference work when using their RNN approach~\cite{PETRUCCI201866} compared to our Transformer model on the same datasets. The first thing to note is that the RNN approach achieves much better results when training with very few samples. It achieves a FA score of over 60\% with only 2k training examples and improves rapidly towards its peak accuracies when training with 5k. Also, the TA score is over 90\% for all experiments, showing that it quickly picks up on the syntactic structure of the data.

\begin{table}[ht]
    \centering
    \begin{tabular}{llll}
        \hline
        Model & FA & ED & TA \\
        \hline
        RNN-2k & 61.1\% & 2.48 & 91.8\% \\
        TF-2k & 0.0\% & 10.2 & 42.6\% \\
        RNN-5k & 84.4\% & 0.6 & 97.5\% \\
        TF-5k & 0.0\% & 9.25 & 51.3\% \\
        RNN-10k & 88.8\% & 0.47 & 98.7 \% \\
        TF-10k & 99.8\% & 0.007 & 99.9\% \\
        RNN-20k & 81.7\% & 0.46 & 98.3\% \\
        TF-20k & \textbf{99.9\%} & \textbf{0.000067} & \textbf{99.9\%} \\
        \hline
    \end{tabular}
    \caption{RNNs vs. Transformers (TF) using the  DKET dataset~\cite{PETRUCCI201866}.}
    \label{tab:dket_results}
\end{table}

In fact, the RNN approach learns the structure much faster than the Transformer does, as we can see that the Transformer performs terribly when training on 2k as well as 5k examples. In neither case the model actually manages to correctly predict a full formula, meaning that the overall FA score is 0\%. In addition to this, the TA scores are around 40-50\% meaning that the predictions are not even close, having to replace over 9 tokens to be correct on average.

However, when the Transformer model gets enough data samples to train on, we see an incredible jump in performance. While the RNN approach improves with $\sim$ 4\% between 5k and 10k examples, we see a literal jump of over 99\% for the Transformer with 10k. When provided with enough data, the Transformer outperforms the RNN's best performance and translates the data almost perfectly. In fact, at its best, when training on 20k examples, it only missed a total of two translations out of the entire set of 30k. This indicates that the syntactic structure of the DKET datasets is easier for the semantically aware Transformer model to capture and translate than for the specified  RNN approach when there is a good amount of training examples to learn from.


\subsection{Atomic on small datasets}
The results of performing experiments on the Atomic datasets quickly show that the Transformer model has a harder time translating these datasets than it had with DKET. Whereas it was almost perfectly translating DKET when it had trained on enough data, we only ever see results on Atomic that are as good as the RNN approach was capable of doing on DKET. This suggests that capturing variables and atoms in first-order logic rules  might be harder for the model than the DL language $\mathcal{ALCQ}$, making Atomic a much more challenging translation task. We observe in Table \ref{tab:atomic_results} that the model increasingly improves (without big jumps) as it is provided with more data.
\begin{table}[t]
    \centering
    \begin{tabular}{l|lll}
        \hline
        Training Dataset & FA & ED & TA \\
        \hline
        2k-Persona & 0.08\% & 2.83 & 84.4\% \\
        5k-Persona & 12.9\% & 1.48 & 93.4\% \\
        10k-Persona & 49.6\% & 0.65 & 97.0\% \\
        20k-Persona & 78.3\% & 0.27 & 98.8\% \\
        \hline
        2k-Mental & 0.03\% & 3.82 & 84.8\% \\
        5k-Mental & 14.2\% & 1.79 & 92.9\% \\
        10k-Mental & 64.0\% & 0.48 & 98.1\% \\
        20k-Mental & \textbf{84.1\%} &\textbf{0.19} & \textbf{99.3\%} \\
        \hline
        2k-Event & 0.0\% & 4.90 & 81.8\% \\
        5k-Event & 3.13\% & 2.64 & 90.2\% \\
        10k-Event & 55.0\% & 0.61 & 97.7\% \\
        20k-Event & 78.9\% & 0.25 & 99.1\% \\
        \hline
        2k-All & 0.0\% & 4.51 & 82.6\% \\
        5k-All & 4.82\% & 2.31 & 91.1\% \\
        10k-All & 35.1\% & 1.01 & 96.1\% \\
        20k-All & 76.3\% & 0.29 & 98.9\% \\
        \hline
    \end{tabular}
    \caption{Results from Persona, Mental-State, Event, and All-included datasets.}
    \label{tab:atomic_results}
\end{table}

%
The TA score is over 80\% for all experiments even when using the smallest training datasets. The reason for this is most likely due to the syntax structure 
of Atomic being simpler than DKET and therefore requiring less training data.
However the
vocabulary of Atomic is roughly three times the size of the one used in DKET, which results in the model struggling more  to create a semantic representation of each word that appears in the dataset,   not achieving the nearly perfect scores as it happened with DKET.

\begin{example}\label{ex:perfect}
We now provide some examples illustrating the translation. Consider the following
    \textbf{oReact if-then relation:} If PersonX kills PersonY's father then PersonY is in grief. 
    In this case, the  correct formula and the predicted one were exactly the same:\
    A x y z ( ( person (x) \& person (y) \& kills (x,z,y) \& father (z) ) $\rightarrow$ grief (y) ). 
\end{example}

As mentioned, Example~\ref{ex:perfect} depicts  a perfectly translated formula, where everything has been captured as expected. All the variables are quantified correctly, the atoms are correct, as well as the     parentheses.
In the next example, on the other hand, we see that the inference word has been switched to a different one, while all the syntactic pieces of the formula are otherwise correct. 

\begin{example}Now consider the following
    \textbf{xAttr if-then relation:} If PersonX loves PersonX'd husband then PersonX is enamored. \\
    \textbf{Correct formula:}\\
    A x z ( ( person (x) \& loves (x,z) \& husband (z) ) $\rightarrow$ \underline{enamored} (x) ) \\
    \textbf{Predicted formula:}\\
    A x z ( ( person (x) \& loves (x,z) \& husband (z) ) $\rightarrow$ \underline{affectionate} (x) )
\end{example}

This mistake might be caused by the fact that ``enamored'' is quite an uncommon word, and often appear in the same contexts as ``affectionate'' which is observed ten times as much in the data. We even see it appear twice just for this specific event, which might have contributed to the incorrect choice of word. The final example that we discuss illustrates the another kind of common issue, where the model makes a syntactic mistake.
\begin{example} Consider the following
    \textbf{xWant if-then relation:}
    If PersonX is really sad then PersonX wants to speak with a friend. \\
    \textbf{Correct formula:}\\
    A x z ( ( person (x) \& is really (x,z) \& sad (z) ) $\rightarrow$ E a b \underline{(} to speak with (x,a)   \& friend (a)   ) ) \\
    \textbf{Predicted formula:}\\
    A x z ( ( person (x) \& is really (x,z) \& sad (z) ) $\rightarrow$ E a b \underline{)} to speak with (x,a)   \& friend (a)   ) )
\end{example}
 In this case, a single parenthesis has been flipped, which in itself is the smallest type of mistake we can find in predictions, but is still considered incorrect like any other mistake. It might seem trivial compared to quantifying all the variables correctly and capturing all the atoms but it shows that sometimes keeping track of all the different parts of the formula patterns at the same time can be difficult.

Interestingly, the categories' results are very similar across the board, except that the mental-state category has a noticeably better result than the other categories when trained on 20k examples. This was unexpected, as one would assume that the Persona category would be the easiest to learn due to only consisting of one relation that always centers around the subject of the event (``PersonX''), and has the shortest formula lengths on average, but this does not seem to be the case. In fact, Persona achieves the same best accuracy at 78\% as the Event category, despite the Event one having the largest set of examples to pick from and the most amount of relations. Thus it seems that the subtleties between the categories do not have a huge impact on the overall results, as all categories seem to improve similarly in all three evaluation metrics as the amount of training samples increases. When sampling from all categories, we also see the same results as doing them separately, showcasing that the model performs well across the board with an overall FA between 76\% to 84\% and a token accuracy at approx. 99\%. This suggests  that the incorrect translations are generally also really close to the correct formula.

This  leads to the overall interpretation of the results. Unsurprisingly,  the biggest factor for performance is the amount of data that the model has to train on. We see that the model is not able to correctly translate at all when only having 2k samples to work with, despite having a high TA, but it improves more and more as the training dataset size increases. From this, we assume that the Transformer's overall performance is hindered by lack of training data and that it  improves as more data is available.

\subsection{Atomic without quantification}

\begin{table}[t]
    \centering
    \begin{tabular}{l|lll}
        \hline
        Training Dataset & FA & ED & TA \\
        \hline
        2k-Persona & 0.06\% & 2.69 & 79.9\% \\
        5k-Persona & 18.2\% & 1.32 & 90.1\% \\
        10k-Persona & 46.1\% & 0.72 & 94.6\% \\
        20k-Persona & 85.7\% & 0.16 & 98.8\% \\
        \hline
        2k-Mental & 0.17\% & 3.53 & 77.6\% \\
        5k-Mental & 16.1\% & 1.67 & 89.2\% \\
        10k-Mental & 61.3\% & 0.51 & 96.6\% \\
        20k-Mental & \textbf{86.7\%} &\textbf{0.15} & \textbf{99.0\%} \\
        \hline
        2k-Event & 0.01\% & 4.32 & 74.4\% \\
        5k-Event & 17.5\% & 1.59 & 90.4\% \\
        10k-Event & 67.9\% & 0.41 & 97.5\% \\
        20k-Event & 84.5\% & 0.19 & 98.9\% \\
        \hline
        2k-All & 0.003\% & 4.21 & 73.8\% \\
        5k-All & 9.44\% & 1.97 & 87.9\% \\
        10k-All & 68.6\% & 0.39 & 97.5\% \\
        20k-All & 80.8\% & 0.23 & 98.6\% \\
        \hline
    \end{tabular}
    \caption{Results from Persona, Mental-State, Event, and All-included datasets without quantification.}
    \label{tab:atomic_results_no_quantification}
\end{table}

Despite the Transformer model's ability to capture  atoms, variables, and to quantify the variables, we do not   expect it to fully comprehend the complex meaning of all of it. The formulas always follow the same pattern of universally quantifying any variable that is found in the body and potentially in the head, while any variable that  occurs exclusively in the head is quantified existentially. This is a result of the algorithm created to make the formulas from the natural language, a deliberate design choice during our process. So, despite there being multiple equivalent ways of translating a sentence correctly, the model has only learned a specific pattern that applies to all examples. It is therefore not   expected to   understand or produce alternate correct interpretations if tested in other logically equivalent but different patterns. 

Thus, you could consider the quantification of the variables  only as a syntactic structure to be learned and the question then becomes how much this impacts the model's translation ability  and if it could achieve  better results when variable quantification is omitted   from the formulas. Their removal 
substantially decreases the 
lengths of the formulas and, thus, maybe we would see the model having an easier time correctly translating,  since fewer total tokens to generate means less potential for mistakes. If the results  increase dramatically, omitting variables may be a more viable solution.
One could run an algorithm on the results that finds all the variables and applies the pattern for quantifying variables as a post-process.

In Table \ref{tab:atomic_results_no_quantification} we see the results of performing the same experiments on the Atomic dataset where the quantification has been removed from the logical formulas. At a first glance, they look very similar. The model performs extremely poorly when trained using the smallest datasets. We can even observe that the overall TA score is worse on the 2k datasets compared to the experiments with the quantification included. This is likely due to the fact that there are fewer ``pattern'' tokens to translate, such as parentheses, $\exists$, and $\forall$ that are always a part of the sequence's variable quantification. 
In the previous experiment, longer sequences can make the model ``seem'' to make more correct  translations, even when making more/the same amount of mistakes as here, when translating shorter sequences, for evaluation metrics which are percentage based. 
This is supported by the fact that the TA score (which is a percentage) is in many cases lower in the experiments here (with shorter sequences) when compared to the experiments with quantification (with longer sequences), despite the  better scores for both   FA and ED.

Interestingly, we notice that the models achieve an overall improvement of approx. 5\% FA in all categories at their peak when trained on 20k training examples. The exception to this case is actually the mental-state category, which was exceptionally good in the quantification experiment as well. Here we only saw a small improvement by a couple of percents in terms of FA, but this is now the new highest score across any experiment. Another interesting observation is that the Event category has a much better result in the 5k training dataset in terms of FA, with an increased score of 14\% despite having the exact same TA as in the equivalent experiment. In the 10k training dataset consisting of all types of relations we also see an FA increase of 33\%, meaning that it can now capture 10k of the 30k validation examples that it was not able to when it had to quantify as well. This was also the worst-scoring experiment using 10k examples with quantification, suggesting that figuring out how to quantify with examples from the different categories was the biggest struggle for the model.

The biggest changes are noticeable in the 5k and 10k datasets, while the differences are much smaller in the 2k and 20k. This suggests that 2k is not enough data to capture the structure of the formulas at all, while at 20k examples the model has enough data to learn, regardless of having the quantification included or not. It matters more in the middle ground, where the model shows more prowess at capturing the atoms when they are its only focus, and it has a harder time when it needs to focus and quantify the variables at the same time. However, this does indeed support the idea that the difficulty of translation is to capture and split up the atoms correctly, as omitting the quantification did not suddenly cause the model to achieve near-perfect accuracy. Instead, we see an overall improvement, suggesting that shorter sequences and more ability to focus on capturing the atoms contributes to better performance.


\subsection{Atomic with all the data for training}

Here we perform experiments using 37k (summing up 2k, 5k, 10k, 20k from before)\footnote{The reasoning for using the datasets split into 2k, 5k, 10k, 20kseen in the previous experiments was to compare them to the reference work using the RNN approach, where achieving good results on a few training samples was something highly valued.}.
We use an 85/15 split, where the model trains on 85\% of the examples and is evaluated on 15\% of them. We performed the experiments on all categories separately as well as the entirety of the dataset with quantification of the variables both included and omitted.
\begin{table}[t]
    \centering
    \begin{tabular}{l|lll}
        \hline
        Training Dataset & FA & ED & TA \\
        \hline
        Persona & 93.6\% & 0.07 & 99.69\% \\
        Mental & 93.7\% & 0.07 & 99.73\% \\
        Event & 93.1\% & 0.07 & 99.72\% \\
        All & \textbf{93.8\%} & \textbf{0.06} & \textbf{99.74\%} \\
        \hline
        Persona-No-Quantification & 94.8\% & 0.05 & 99.59\% \\
        Mental-No-Quantification & 94.7\% & 0.06 & 99.60\% \\
        Event-No-Quantification & 93.3\% & 0.07 & 99.57\% \\
        All-No-Quantification & 93.8\% & 0.06 & 99.58 \\
        \hline
    \end{tabular}
    \caption{Results with all the data,   comparing with and without quantification.}
    \label{tab:atomic_results_full}
\end{table}
From Table \ref{tab:atomic_results_full} we can immediately observe that the model performs better across the board than any of the experiments on the smaller datasets did. The variation is also incredibly low, where we see a difference of less than 2\% from best to worst in terms of FA. The ED is also very low, averaging less than half of the best score from the smaller dataset experiments, where the mental state on 20k training examples had the best of $0.15$ while now all are at $0.07$ and below. The TA score is now also always above 99.5\% meaning that only $1$ out of $200$ tokens are incorrect in the translation, suggesting that even the incorrect formulas are extremely close to the correct.
We see that now that the model has such a larger amount of data to train on, it no longer matters if the quantification is omitted or not. 
The edge of the  results with omitted variables is almost insignificant in the last case.

\section{Conclusion}

Logical reasoning is a challenging task even
for the most advanced language models and
increasing the amount of data does not 
really solve the issue in general~\cite{zhang_paradox_2022}.
In this work, we see reasoners as the most reliable tools for performing logical reasoning and focus on translating natural language into the logic format expected 
by such tools. 
We devised an algorithm for creating a dataset with statements from the Atomic knowledge bank and their respective translations into first-order logic rules. 
Then, we explored the capability of Transformers
to translate statements expressing rules and definitory sentences
into first-order logic rules and $\mathcal{ALCQ}$  concept inclusions (respectively).

%
%
%
 \bibliographystyle{splncs04}
 \bibliography{ref.bib}
 \end{document}